\documentclass[conference]{IEEEtran}  
\IEEEoverridecommandlockouts                                                          

\usepackage{cite}

\usepackage{amsmath,amssymb,amsfonts}

\usepackage{gensymb}

\usepackage{graphicx}
\usepackage{textcomp}

\usepackage{xcolor}

\usepackage{verbatim}

\usepackage{afterpage}

\usepackage{cite}

\usepackage{overpic}

\usepackage{psfrag}

\usepackage{latexsym}

\usepackage{bm}

\usepackage{cases}

\usepackage{array}

\usepackage{fancyhdr}

\usepackage{setspace}

\usepackage{subfigure}

\usepackage{url}

\usepackage{multirow}

\usepackage{epstopdf}

\usepackage{epsfig}

\usepackage{fancybox}

\usepackage{textcomp}

\usepackage[ruled,linesnumbered]{algorithm2e}

\usepackage{geometry}
\DeclareMathOperator*{\argminA}{arg\,min} 
\newcommand{\x}[0]{\mathbf{x}}
\newcommand{\T}[0]{\mathbf{T}}
\newcommand{\ranging}[0]{\mathbf{r}}
\newcommand{\m}[0]{\mathbf{m}}
\newcommand{\M}[0]{\mathbf{M}}
\newcommand{\Z}[0]{\mathbf{Z}}
\newcommand{\z}[0]{\mathbf{z}}
\newcommand{\e}[0]{\mathbf{e}}

\begin{document}

\newgeometry{left=54pt,right=54pt,top=72pt,bottom=54pt}

\title{Relative Localization of Mobile Robots with Multiple Ultra-WideBand Ranging Measurements}

\author{Zhiqiang Cao, Ran Liu, Chau Yuen, Achala Athukorala, Benny Kai Kiat Ng, \\
Muraleetharan Mathanraj, and U-Xuan Tan
\thanks{This work is supported by National Key R\&D Program of China (2019YFB1310805). Z. Cao, R. Liu, C. Yuen, A. Athukorala, B. K. K. Ng, U-X. Tan are with the Singapore University of Technology and Design, 8 Somapah Road, Singapore 487372. M. Mathanraj is with the University of Jaffna, Jaffna, Sri Lanka 40000. Z. Cao and R. Liu are also with the Southwest University of Science and Technology, Mianyang, Sichuan, China 621010. Corresponding author: \tt\small  ran\_liu@sutd.edu.sg. }
}

\maketitle

\begin{abstract}
Relative localization between autonomous robots without infrastructure is crucial to achieve their navigation, path planning, and formation in many applications, such as emergency response, where acquiring a prior knowledge of the environment is not possible.
The traditional Ultra-WideBand (UWB)-based approach provides a good estimation of the distance between the robots, 
but obtaining the relative pose (including the displacement and orientation) remains challenging.
We propose an approach to estimate the relative pose between a group of robots by equipping each robot with multiple UWB ranging nodes. 
We determine the pose between two robots by minimizing the residual error of the ranging measurements from all UWB nodes. 
To improve the localization accuracy, we propose to utilize the odometry constraints through a sliding window-based optimization. 
The optimized pose is then fused with the odometry in a particle filtering for pose tracking among a group of mobile robots.
We have conducted extensive experiments to validate the effectiveness of the proposed approach.
\end{abstract}

\section{Introduction}
\label{sec:introduction}
Localization is essential for robots to achieve true automatic movement \cite{ref1}.
The literature shows a number of mature techniques and implementations for localization given a known infrastructure (i.e., map of the environment or distribution of the landmarks) \cite{ref2}\cite{ref3}. 
One example is the use of GPS to provide meter-level positioning accuracy by trilaterating the signals from at least four satellites in outdoor environments, which is not suitable for indoor positioning due to the block of the satellite signals from buildings \cite{ref5}. 

\begin{figure}
\centerline{\includegraphics[width=8cm]{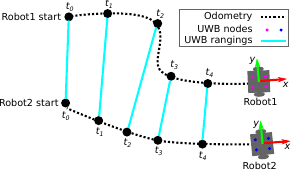}}
\caption{Overview of the proposed approach of relative localization of multiple robots (two robots as an example). 
Each robot carries a set of UWB nodes for ranging. 
Our goal is to achieve the relative localization between a group of robots based on the fusion of UWB ranging and odometry measurements.
}
\label{fig:overview}
\vspace{-0.2cm}
\end{figure}

However, a prior knowledge of the environment is not always available. 
For example, the emergency response requires a team of robots to explore an unknown site, where the existing infrastructure might be damaged \cite{ref10}. 
In this case, knowing the relative position between the robots is vital to perform the exploration task effectively. 
A number of sensors, for example visual and LiDAR, can be used to achieve the relative localization. 
These approaches provide a good localization accuracy when the robots are within the field of view, 
but it is challenging to deal with occlusions and the algorithms often require a lot of computational resources. 
Radios (for example WiFi and UWB) are widely used for relative localization due to the detection of signals without line-of-sight \cite{ref9}.
For example, UWB sensors provide centimeter-level accuracy and offers a maximum reading range up to 100 meters.
Recent research shows a growing deployment of UWB for the positioning in robotics community, due to its low cost, small size, and low power consumption. 

Although a precise distance between the robots can be obtained by the UWB, the lack of bearing information makes it difficult to be applied in industrial environments \cite{ref12}. 
Acquiring the bearing is critical for autonomous robots to perform various tasks, for example navigation, docking, and formation control. 
The traditional Angle-of-Arrival-based solutions require an expensive setup to determine phase shift from different antennas, 
which are not suitable for robotics applications. 
Therefore, this paper proposes an approach to estimate the pose (position and orientation) between the robots using a set of ranging measurements from UWBs. 
In particular, we equip each robot with a number of UWB nodes as shown in Figure \ref{fig:overview}. 
The installed positions of UWB nodes on a specific robot are known in advanced. 
We achieve the relative pose estimation by minimizing the residual error of the ranging measurements from the multiple UWB nodes installed on different robots.

The UWB ranging is susceptible to multipath phenomenon in indoor environments \cite{ref14}. 
This challenges the accurate pose estimation using the ranging difference from the UWBs, 
in particular when the distance between the UWB nodes on the
\restoregeometry
same robot is short. 
Odometry determines the location of a robot with dead reckoning\cite{ref15} based on wheel encoder or Inertial Measurement Unit (IMU).
Although the odometry will drift for a long term run, the estimation of odometry in a short time is accurate. 
Hence in this paper, we propose to refine the pose estimation by 
considering the odometry as constraints through a sliding window-based optimization. 
Finally, to track the pose of the moving robot, we propose to use the particle filtering (PF) to fuse the optimized pose and the odometry measurements.  
We evaluated the performance of the proposed approach by a series of experiments. 
The contributions of this paper are summarized as follows:
\begin{enumerate}
\item We propose a method to realize the relative pose estimation between mobile robots through non-linear optimization of the UWB ranging measurements from multiple UWB nodes mounted on different robots.
\item We incorporate odometry measurements as constraints to improve the accuracy of pose estimation based on a sliding window-based optimization.
\item We fuse the odometry and the optimized pose obtained from the sliding window-based approach using the particle filtering to track the pose of a group of moving robots.
\item We have performed extensive experiments to evaluate the performance of the proposed approach. The results show 
that we achieved a positioning accuracy of 0.312$m$ in translation and 4.903$^\circ $ in rotation by fusing the UWB and odometry measurements when three robots are moving in the environment with a size of 6m$\times$12m.
\end{enumerate}

The remainder of this paper is organized as follows: Section\,II introduces the related work. 
Section\,III describes the approaches for relative pose estimation using UWB rangings, pose optimization with additional odometry constraints, and the particle filtering for pose tracking. 
Extensive experiments to validate the proposed approach are conducted in Section\,IV. Finally, the conclusion and future work are given in Section\,V.

\section{Related Work}
\label{sec:related_work} 
Despite GPS is widely used in outdoor environment, it cannot be applied in indoor scenarios, as satellite signals are easily reflected and diffracted by city buildings \cite{ref20}. A large number of researchers focused on the localization in GPS-denied environments with different sensor techniques and implementations. 
Visual and LiDAR are two widely used sensors that provide high positioning accuracy, 
with a number of applications in the industry. 
Authors in \cite{ref32} proposed to use self-optimized-ordered visual vocabulary to find deep connections between feature clusters and physical locations.
Authors in \cite{ref33} proposed an approach that leverages visual information captured by surveillance cameras and smart devices to deliver accurate location information.
Authors in \cite{ref34} proposed to use full information maximum likelihood optimal estimation to improve the positioning accuracy of a device equipped with LiDAR.

In many scenarios, however, the prior knowledge about the infrastructure is not known. 
Therefore, relative localization without any infrastructure is crucial for the robot to perform navigation and formation control.
Due to its low cost and precise ranging accuracy \cite{ref24}, UWB has been used for localization in many industrial applications. 
Another advantage is that UWB works well in non-line-of-sight environments, due to its good capability to penetrate through a variety of materials including walls, metals, and liquids.
Authors in \cite{ref16} used four transceivers as base stations to improve the robustness and precision of the positioning system. 
But the orientation of the object cannot be estimated via this method.
Authors in \cite{ref26} proposed a waveform division multiple access scheme to enhance the positioning accuracy in multi-user applications.
In order to improve the performence of relative localization,
authors in \cite{ref11} proposed a visual-inertial-UWB fusion framework for relative state estimation.
Meanwhile, authors in \cite{ref33} introduced a decentralized omnidirectional visual-inertial-UWB state estimation system to solve the issues of observability, complicated initialization, and insufficient accuracy.

Odometry plays an essential role for localization, 
as it can provide accurate pose estimation for a mobile robot in a short period \cite{ref19}. 
Therefore, many researchers focus on the fusion of odometry with other sources of sensors to improve the localization accuracy. 
Authors in \cite{ref27} proposed an indirect cooperative relative localization method to estimate the position of a group of unmanned aerial vehicles based on distance and IMU displacement measurements. 
However, they only discussed the problem of direct relative positioning to a static UAV. 
Authors in \cite{ref28} proposed an approach to combine IMU inertial and UWB ranging measurement for relative positioning between multiple mobile users without the knowledge of the infrastructure. 
But they assume the initial poses of the users are known. 
Authors in \cite{ref29} proposed a method to fuse IMU, magnetometers, peer-to-peer ranging, and downward looking cameras to provide high-precise estimation of UAVs.
However, this method has high computational complexity and it is not suitable for non-line-of-sight applications. 

\section{Relative Pose Estimation}
In this section, we first formulate the problem of the relative positioning with multiple UWB rangings to localize a group of robots. 
Then, we show the details of our proposed relative positioning approach, which consists of three modules. 
First, we achieve the relative pose estimation using UWB ranging measurements from multiple UWB ranging measurements.
Second, we refine the pose estimation by incorporating odometry constraints through a sliding window-based optimization technique.
Third, a particle filtering is used to fuse the odometry and optimized pose for the tracking of a group of robots.

\subsection{Problem Formulation}
\label{problem_formulation}
Figure \ref{fig:overview} shows a scenario where a group of robots needs to perform relative localization without any given infrastructure. 
Each mobile robot carries a set of UWB nodes to provide ranging information from neighbor robots in communication range. 
Additionally, each robot offers odometry measurement, which provides its relative movement between adjacent timestamps. 
The goal is to achieve relative localization between multiple mobile robots through the UWB ranging and odometry measurements without a prior knowledge about the infrastructure. 
Our approach works in a centralized fashion, which requires all robots to send the measurements (i.e., odometry and UWB) to a server for pose estimation. 

Formally, lets denote the pose of the robot $i$ ($i \in [1:N]$) at time $t$ as ${{\x}_i^{(t)}} = {[x_i^{(t)},y_i^{(t)},\theta_i^{(t)}]}$, 
where $x_i^{(t)}$ and $y_i^{(t)}$ represent its 2D position and $\theta_i^{(t)}$ denotes its orientation at time $t$.
The odometry of each robot provides its relative movement ${\m}_{i}^t=[\Delta {x}_i^t, \Delta {y}_i^t, \Delta {\theta}_i^t]$ at time $t$. 
We denote the relative position of UWB $k$ on robot $i$ as $C_i^k, k \in [1: K]$, where $K$ is the total number of UWB nodes on robot $i$. 
Similarly, the position of UWB $l$ on robot $j$ is denoted as $C_j^l, l \in [1: L]$, where $L$ is the total number of UWB nodes on robot $j$. 
$r_{i^k,j^l}^{(t)}$ represents the UWB ranging measurement between UWB node $k$ on robot $i$ and UWB node $l$ on robot $j$.
At time $t$, robot $i$ will receive $L \times K$ UWB ranging measurements from robot $j$, which is denoted as ${\ranging}_{i,j}^t=\{r_{i^k,j^l}^{(t)}\}_{k,l=1}^{K, L}$. 
$N_{i}^{(t)}$ represents the set of robots that are in range of robot $i$ at time $t$.
Therefore, we use $\z_i^{(t)}=\{\ranging_{i, j}^{(t)}\}$ ($j \in N_{i}^{(t)})$ to denote the total UWB ranging measurements received by robot $i$ at time $t$. We aim to determine the relative pose of robot $i$ with respect to robot $j$ at time $t$, 
which includes the estimation of 2D relative position and orientation. 

\subsection{Relative Pose Estimation based on UWB Rangings}
\label{pose_est_ranging}
The estimation of the relative pose between robot $i$ and robot $j$ can be achieved by finding the best configuration of the pose through minimizing the residual error of the UWB ranging measurements:
\begin{equation}
 \begin{split}
\overline{\T}_{i,j}^{t}= \argminA_{\x=(x,y,\theta)} \sum_{l=1,k=1}^{l=L, k=K} (r_{i^k,j^l}^{(t)}-d(\x,C_i^k,C_j^l))^2
\label{eq:optimization}
 \end{split}
\end{equation}
where the function $d(.)$ computes the distance between UWB node $k$ on robot $i$ and UWB node $j$ on robot $k$ given a pose $\x$.
The minimization in Equation \ref{eq:optimization} is considered as a non-linear optimization problem, which is solved by general graph optimization (g2o) \cite{ref18} in this paper.
In particular, the pose to be estimated is denoted as the node in the graph and the constraints are represented by the UWB ranging measurements. The algorithm turns out to find the best configuration of the node (i.e., relative pose $\overline{\T}_{i,j}^{t}$ to be estimated) 
to satisfy the UWB ranging constraints (i.e., measurements) based on maximum likelihood estimation.  

\subsection{Pose Optimization by Adding Odometry Constraints}
\label{pose_est_g2o}
Due to the limitation of the size of the robot, the distance between the node on a robot is set to be small, 
which requires a high UWB ranging accuracy to guarantee a good relative pose estimation. 
This is particularly challenging for indoor environment, 
as the ranging of UWB is highly impacted by the non-line-of-sight signal propagation.
Therefore, we consider to apply the odometry constraints to improve the accuracy of pose estimation. 
Our solution is to perform pose graph optimization using a sliding window-based technique by considering the following two types of constraints: namely odometry constraints ${\m}_{i}^t$ and pose constraints $\overline{\T}_{i,j}^{t}$ obtained from the UWB rangings (see Section \ref{pose_est_ranging}): 
\begin{equation}
 \begin{split}
&\argminA_{\x} \sum_{i=1}^{N} \sum_{t'=t-w}^{t} \underbrace{ {\e} ({\x_i^{t'-1},\x_i^{t'}, \m_i^{t'} })^T \Omega_{i}^{t'}  {\e} ({\x_i^{t'-1},\x_i^{t'}, \m_i^{t'} }) }_{\text{Odometry-based constraint}}\\
&+\sum_{i=1}^{N} \sum_{j \neq i}^{N} \sum_{t'=t-w}^{t} \underbrace{ {\e} ({\x_i^{t'},\x_j^{t'}, \overline{\T}_{i,j}^{t'} })^T \Omega_{i,j}^{t'}  {\e} ({\x_i^{t'},\x_j^{t'}, \overline{\T}_{i,j}^{t'} }) }_{\text{Pose constraint using UWB ranging (Sect.\,\ref{pose_est_ranging})}}
 \label{eq:graph_optimization}
 \end{split}
\end{equation}
where $w$ represents the size of the sliding window used for optimization. 
$\e(\cdot)$ denotes the error function which is computed based on the given poses and the constraints inferred from the observations (i.e.,
odometry $\m_i^{t}$ and pose estimation $\overline{\T}_{i,j}^{t}$ obtained from UWB ranging measurements in Section \ref{pose_est_ranging}). 
Both constraints are represented as a 3$\times$1 vector, which includes the 2D displacement and the orientation.
Constraints are additionally parameterized with a certain degree of uncertainty, which is denoted as the information matrix (i.e., $\Omega_{i}^{t'}$ and $\Omega_{i,j}^{t'}$) in Equation \ref{eq:graph_optimization}. 
As a result, 
we obtain an optimized pose estimation $\hat{\T}_{i,j}^{t}$ at time $t$ by incorporating odometry constraints within a time window $w$.
The optimized pose $\hat{\T}_{i,j}^{t}$ is then passed into a particle filtering for the tracking of the poses of the robot with a combination of odometry, see Section \ref{pose_est_pf}.

\subsection{Pose Estimation with Particle Filtering}
\label{pose_est_pf}
When robot $i$ moves in the environment, a series of odometry measurements (i.e., ${\M}=\{\m_i^{(1)},\m_i^{(2)},...,\m_i^{(t)} \}$) and UWB ranging measurements 
(i.e., ${\Z}=\{\z_i^{(1)}, \z_i^{(2)}, ..., \z_i^{(t)} \}$) are recorded. 
The goal is to estimate the joint posterior probability $P({{\x}_i^{(t)}}\left| {\M}\right.,{\Z})$ based on odometry measurements $\M$ and UWB ranging measurements $\Z$ in a global frame. 
Instead of using the ranging $\z_i^{(t)}$ as the observations, we use the optimized pose $\hat{\T}_{i,j}^{t}$ obtained from Section \ref{pose_est_g2o} for updating the belief of posterior probability.
Given the fact that odometry and UWB ranging measurements are independent, 
$P({{\x}_i^{(t)}}\left| {\M}\right.,{\Z})$ can be expressed as follows:
\begin{equation}
 \begin{split}
&P({{\x}_i^{(t)}}\left| {\M} \right.,{\Z})\\
&= {\eta_t} \cdot P({{\x}_i^{(t)}}|{{\x}_i^{(t-1)}},{\M}) \cdot P({\Z}|{{\x}_i^{(t)}}) \cdot P({{\x}_i^{(t-1)}}|{\M},{\Z})\\
&= {\eta_t} \cdot P({{\x}_i^{(t)}}|{{\x}_i^{(t-1)}},\m_{i}^{(t)}) \cdot \mathop \prod \limits_{j \in  N_{i}^{(t)}} \mathop P(\hat{\T}_{i,j}^{t}|{{\x}_i^{(t)}}, \x_j^{(t)})\\
&~~~~\cdot P({{\x}_i^{(t-1)}}\left| {\M} \right.,{\Z})
 \label{eq:particle filtering}
 \end{split}
\end{equation}
where ${\eta _t}$ denotes normalization coefficient, which ensures the sum of total probability is one. $P({{\x}_i^{(t)}}|{{\x}_i^{(t-1)}},\m_{i}^{(t)})$ represents the motion model of the mobile robot, which predicts the current state of the robot ${{\x}_i^{(t)}}$ (i.e., position and orientation) at time $t$ based on the previous state $\x_i^{(t-1)}$ and odometry $\m_i^{(t)}$. $P(\hat{\T}_{i,j}^{t}|{{\x}_i^{(t)}},\x_j^{(t)})$ represents the UWB observations model, which gives the likelihood of obtaining a optimized pose $\hat{\T}_{i,j}^{t}$ given the current states of ${{\x}_i^{(t)}}$ and ${{\x}_j^{(t)}}$. 
We use the particle filtering \cite{ref28}\cite{ref30} as the implementation, due to its advantage in dealing with non-linear and non-Gaussian systems when compared with the Kalman filtering. 
The algorithm uses a number of particles to estimate the posterior probability distribution of the robot pose, 
i.e., $\{\x_i^{(s,t)}, w_i^{(s,t)}\}_{s=1}^{S}$, where $S$ denotes the number of particles. 
Each particle carries two kinds of information, namely the pose $\x_i^{(s,t)}$ (i.e., position $x_i^{(s,t)}$, $y_i^{(s,t)}$, and orientation $\theta_i^{(s,t)}$) and the weight $w_k^{(s,t)}$. 
The particle filtering is carried out with two steps, namely prediction and update, which will be described in the rest of this section. 
An overview of the approach is shown in Algorithm \ref{particle_filter_algo}.

\subsubsection{Prediction}
According to the motion model $P({{\x}_i^{(t)}}|{{\x}_i^{(t - 1)}},{\M})$, we predict the pose 
(i.e., position and orientation) of a particle according to: 
\begin{equation}
 \begin{split}
&x_i^{(s,t)} = x_i^{(s,t-1)} + \Delta {d_i^{(t)}} \cdot \cos ({\theta_i ^{(s, t - 1)}}) + {\cal N}(0,\sigma _d^2)\\
&y_i^{(s,t)} = y_i^{(s,t-1)} + \Delta {d_i^{(t)}} \cdot \sin ({\theta_i ^{(s, t - 1)}}) + {\cal N}(0,\sigma _d^2)\\
&\theta_i^{(s,t)} = \theta_i^{(s,t - 1)} + \Delta {\theta_i ^{(t)}} + {\cal N}(0,\sigma _\theta ^2)
\label{eq:prediction}
 \end{split}
\end{equation}
where $\Delta {d_i^{(t)}}$ represents the moving distance of the robot between two adjacent timestamps (i.e., $\Delta {d_i^{(t)}} = \sqrt {{{(\Delta {x_i^{(t)}})}^2} + {{(\Delta {y_i^{(t)}})}^2}} $). 
${\cal N}(0,\sigma _d^2)$ and ${\cal N}(0,\sigma _\theta ^2)$ denote that Gaussian noise with the standard deviations of $\sigma _d$ and $\sigma_\theta$, which are applied to the displacement and orientation of the robot movement, respectively.

\subsubsection{Update}
In this step, the weight of each particle is updated based on the optimized pose from Section \ref{pose_est_g2o}. 
The likelihood of obtaining a pose estimation $\hat{\T}_{i,j}^{t}$ is computed as:

\begin{equation}
 \begin{split}
P(\hat{\T}_{i,j}^{t}|{{\x}_i^{(t)}},\x_j^{(t)})=\frac{1}{{\sqrt {2\pi} {\lambda_d \lambda_\theta} }}\exp ( -\frac{1}{2}d^2(\hat{\T}_{i,j}^{t},{{\x}_i^{(t)}},{{\x}_j^{(t)}}) )
\label{eq:update}
 \end{split}
\end{equation}
where $d^2(\cdot)$ assesses the translational and rotational displacements given the optimized pose $\hat{\T}_{i,j}^{t}$ and the relative pose between ${\x}_i^{(t)}$ and ${\x}_i^{(t)}$. ${{\lambda_d}}$ and ${{\lambda_\theta}}$ represent standard deviations of the optimized pose $\hat{\T}_{i,j}^{t}$ in displacement and orientation, respectively. In particular, $d^2(\cdot)$ is computed as:
\begin{equation}
 \begin{split}
d^2(\cdot)=\frac{(\hat{x}_{i,j}^{t}-{x}_{i,j}^{t})^2}{\lambda_d} + \frac{(\hat{y}_{i,j}^{t}-{y}_{i,j}^{t})^2}{\lambda_d} + \frac{(\hat{\theta}_{i,j}^{t}-{\theta}_{i,j}^{t})^2}{\lambda_\theta}
\label{eq:update1}
 \end{split}
\end{equation}
After the update step, the resample will be performed to generate a new particle set as a replacement the previous particle set, which is critical to avoid degeneration of the particles. 
In principle, the higher the weight of the particles, the greater the probability of being selected during the resampling process. 

\begin{algorithm}

\label{particle_filter_algo}

\SetKw{KwWith}{with}

\SetKw{KwEach}{each}

  \SetKwData{Left}{left}\SetKwData{This}{this}\SetKwData{Up}{up}
  \SetKwFunction{Union}{Union}\SetKwFunction{FindCompress}{FindCompress}
  \SetKwInOut{Input}{input}\SetKwInOut{Output}{output}
\KwData{Previous state $\{\x_i^{(t-1)}\}_{i=1}^{N}$, odometry $\{\m_i^{(t)}\}_{i=1}^{N}$, and UWB ranging $\{\z_i^{(t)}\}_{i=1}^{N}$}

\KwResult{Relative pose estimation at $t$: $\{\x_i^{(t)}\}_{i=1}^{N}$}

\caption{The proposed approach for relative localization between a group of robots}

\tcp{Pose estimation based on UWB ranging (Sect.\,\ref{pose_est_ranging})}
  \For{$i\leftarrow 1$ \KwTo $N$}{
    \For{$j\leftarrow 1$ \KwTo $N$ \KwWith $j \neq i$ }{    
   $\rhd$ Compute the relative pose $\overline{\T}_{i,j}^{t}$ between $i$ and $j$ at time $t$ according to Equation \ref{eq:optimization}\
    }
  }
\tcp{Pose optimization based on odometry constraints (Sect.\,\ref{pose_est_g2o})}
$\rhd$ Pose graph optimization according to Equation \ref{eq:graph_optimization} \\
$\rhd$ Compute the optimized relative pose $\hat{\T}_{i,j}^{t}$ \\
\tcp{Particle filtering for sensor fusion (Sect.\,\ref{pose_est_pf})}
\For{$i\leftarrow 1$ \KwTo $N$}{
$\rhd$ Predict $\x_i^{(t)}$ according to Equation \ref{eq:prediction} given odometry $\m_i^{(t)}$\\

    \For{$j\leftarrow 1$ \KwTo $N$ \KwWith $j \neq i$ }{    
$\rhd$ Update the particle filtering according to Equation \ref{eq:update} based on $\overline{\T}_{i,j}^{t}$\\
    }
\tcp{Resampling}
$\rhd$ Draw new particle sets according to the weights
}

\end{algorithm}

\section{Experimental Results}

\subsection{Experimental Settings}

\begin{figure}

\centering

 \subfigure[Experimental platform]{


    \includegraphics[width=3.0in]{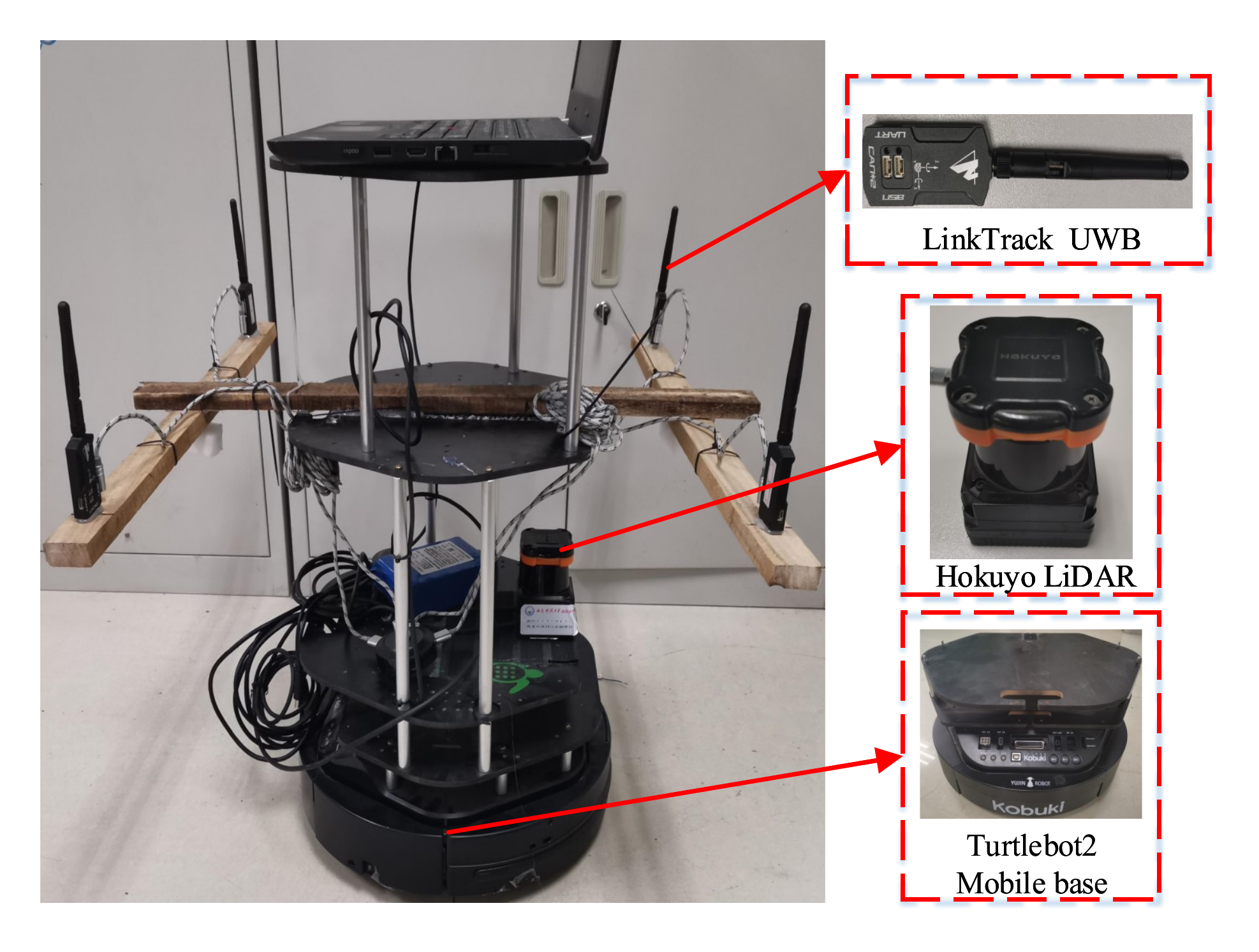}

}

\subfigure[First test case: one static and one mobile robot ]{


    \includegraphics[width=2.8in]{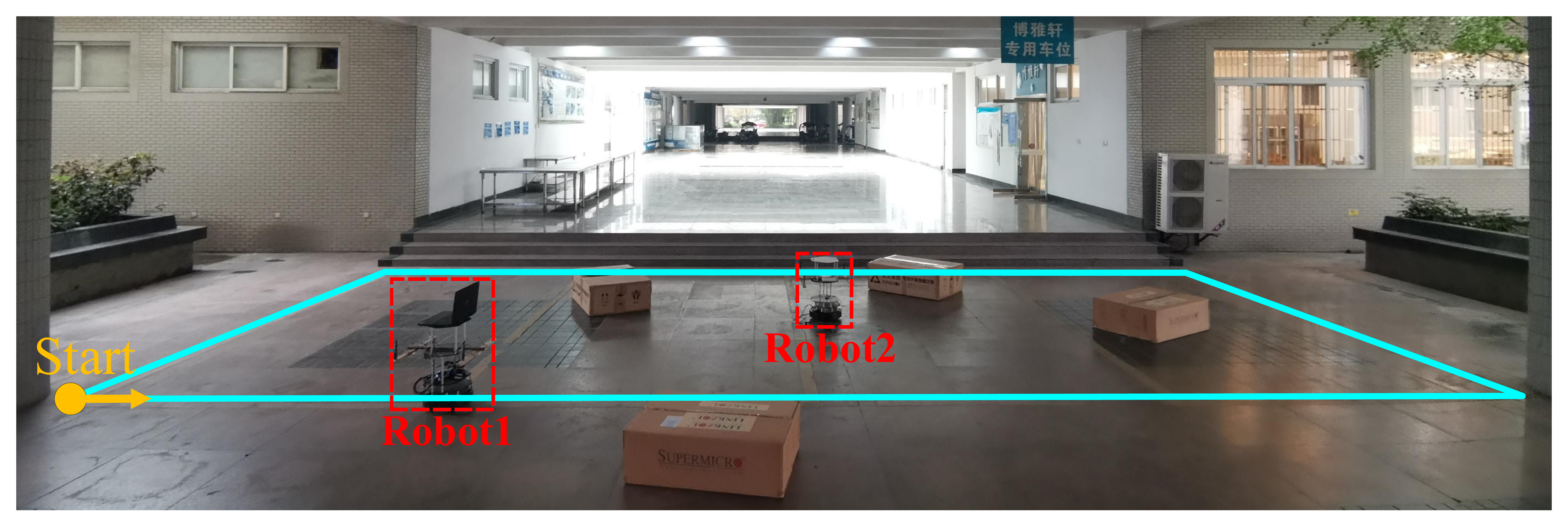}

}

\subfigure[Second test case: three mobile robots]{


    \includegraphics[width=2.9in]{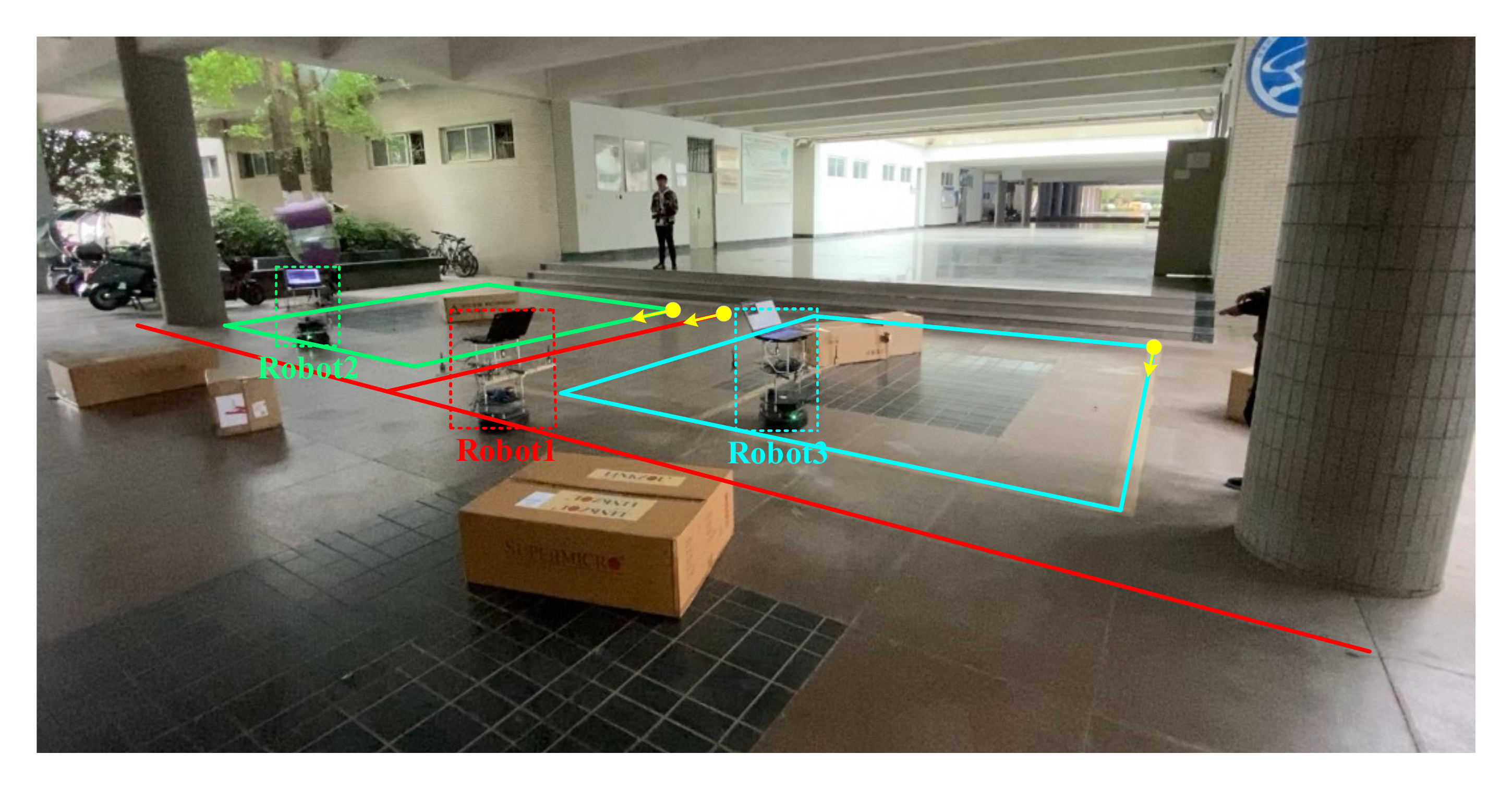}

}

\caption{Overview of the experimental setup. (a) Robot platform and sensors carried by the robot; (b) Robot1 is moving along a rectangular path and robot2 is stationary; (c) Three robots are controlled to move simultaneously. Robot1 moves along the T path, Robot2 and Robot3 move along different rectangular paths.}

\label{experimental_setup} 

\end{figure}

In this section, we present extensive experimental evaluations to demonstrate the proposed approaches with different settings in an indoor environment. In our setup, each robot carried four UWB nodes (LinkTrack), which are able to provide a maximum reading range up to 100 meters.
 These UWB nodes are placed in a square configuration on the robot. The sampling rate of UWB is set to 50Hz. The robot outputs the odometry measurements with a frequency of 20Hz.
 To obtain the ground truth, Hokuyo LiDARs are installed on all robots to perform adaptive Monte Carlo localization (AMCL) \cite{ref31} given a map created by GMapping \cite{gmapping}.
 An overview of the experimental setup is shown in Figure \ref{experimental_setup}(a).

In order to analyze the stability and effectiveness of the proposed method, we designed two test cases, as shown in Figure \ref{experimental_setup}. For the first test case (see Figure \ref{experimental_setup}(b)), one robot (robot2) is stationary and one robot (robot1) is moving along a rectangular path with a size of $7$m$\times6$m.
 For the second test case (see Figure \ref{experimental_setup}(c)), we have three robots moving along different paths simultaneously. Two robots are moving along two different square paths with a size of $5$m$\times5$m and one is moving along a T-shape path with a size of $6$m$\times12$m.
 During the experiment, the maximum velocity of the robot was set to 0.2m/s.
 The relative pose (including the translation and rotation) at a timestamp is compared between the ground truth and our estimation. We then compute the mean squared error (MSE) in translation and rotation to evaluate the positioning accuracy of our approach. In the next sections, we describe the experimental results of the two test cases. In both experiments, we set the number of particles $S=500$, $\sigma_d=0.1$, and $\sigma_\theta=0.05$ for the particle filtering.
 In addition, we use $\lambda_d=1.0$ and $\lambda_\theta=0.1$ to update the weights in the particle filtering.
We refer the readers to \cite{ref10} for a detailed settings of these parameters in the particle filtering.

\begin{figure}

\centering

 \subfigure[Estimated trajectories of robot1 while robot2 remains static]{


    \includegraphics[width=3.0in]{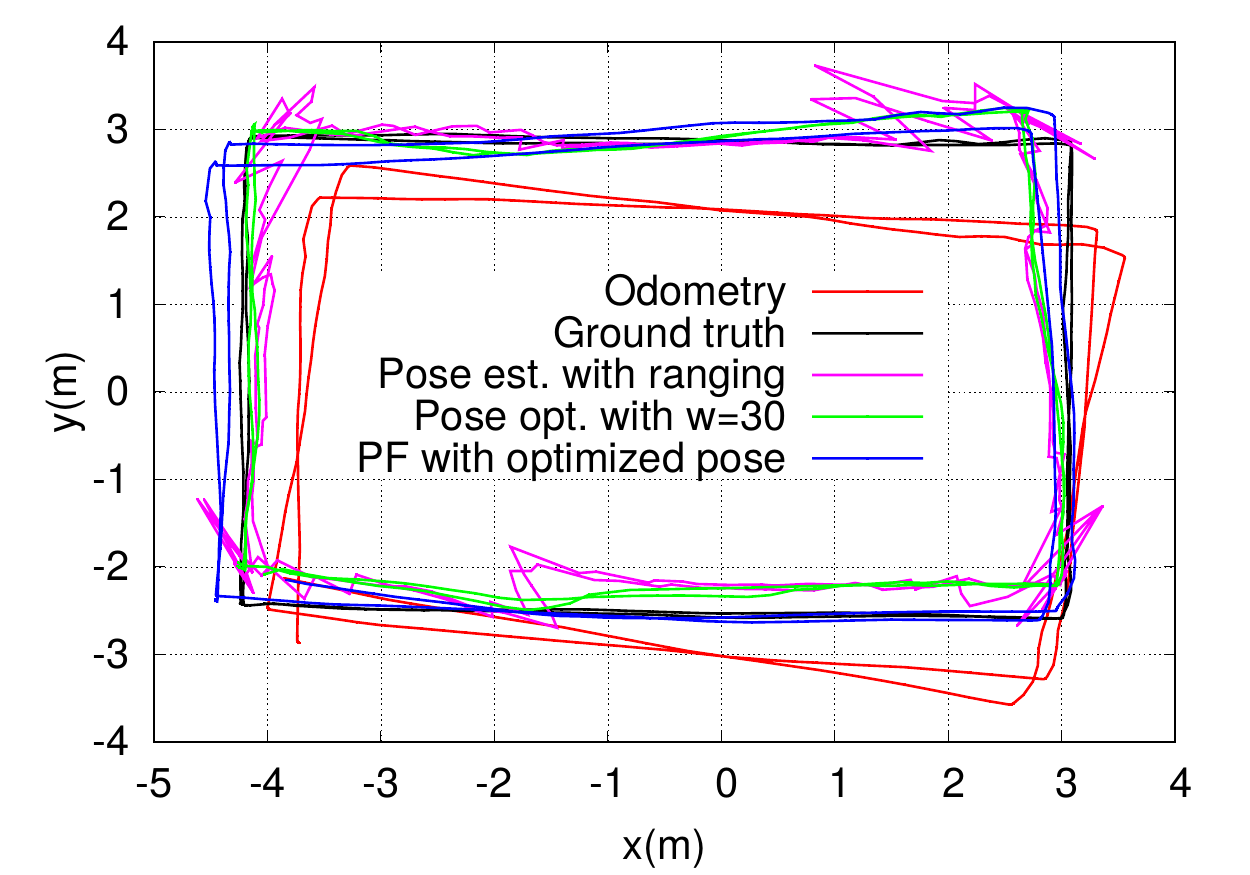}

}

\subfigure[Translational error]{


    \includegraphics[width=3.0in]{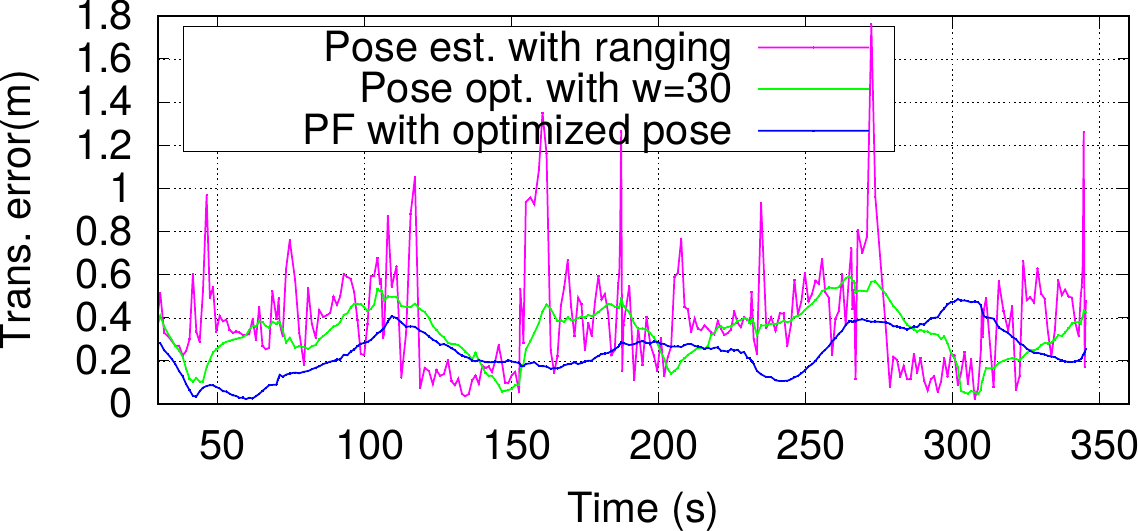}

}

\subfigure[Rotational error]{


    \includegraphics[width=3.0in]{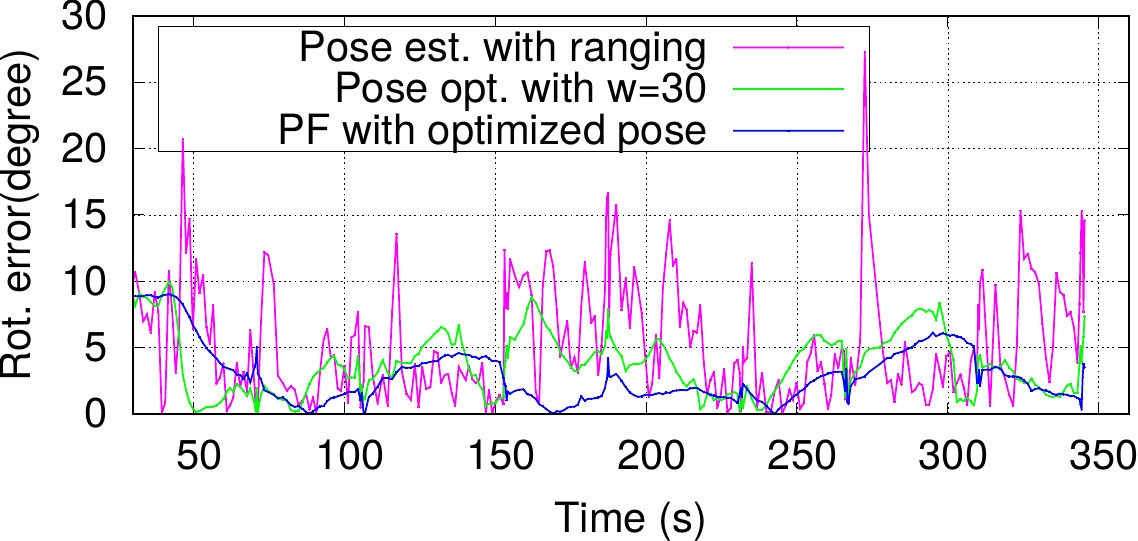}

}

\caption{Experimental evaluation of the first test case. (a) Estimated trajectories of robot1 under different approaches while robot2 remains static; (b) Translational error (in meters) at different timestamps under different approaches; (c) Rotational error (in degrees) at different timestamps under different approaches. }

\label{two_robots_track} 
\end{figure}

\begin{table*}[]
\centering
\caption{Evaluation of the first test case: the translational (in meters) and rotational (in degrees) error of robot1 with respect to robot2 under different configurations of UWB nodes and the performance under different approaches 
i.e., relative pose estimation by UWB ranging (Section \ref{pose_est_ranging}), pose optimization by additional odometry constraints (Section \ref{pose_est_g2o}), the particle filtering that fuses the odometry and UWB ranging\cite{ref10}, and the particle filtering that fuses the odometry and the optimized pose (Section \ref{pose_est_pf}).}

\label{table_static}
\begin{tabular}{|l|l|l|l|l|l|l|}
	\hline
	\multicolumn{1}{|c|}{\multirow{3}{*}{Approaches}} & \multicolumn{6}{c|}{Distance between UWB nodes on the robot}                                                                                                                                                                                                                                                                                                                                                                                                                                        \\ \cline{2-7} 
	\multicolumn{1}{|c|}{}                            & \multicolumn{2}{c|}{0.3m}                                                                                                                                       & \multicolumn{2}{c|}{0.5m}                                                                                                                                       & \multicolumn{2}{c|}{0.7m}                                                                                                                                       \\ \cline{2-7} 
	\multicolumn{1}{|c|}{}                            & \multicolumn{1}{c|}{\begin{tabular}[c]{@{}c@{}}Trans.\\ error (m)\end{tabular}} & \multicolumn{1}{c|}{\begin{tabular}[c]{@{}c@{}}Rot.\\ error ($^\circ$)\end{tabular}} & \multicolumn{1}{c|}{\begin{tabular}[c]{@{}c@{}}Trans.\\ error (m)\end{tabular}} & \multicolumn{1}{c|}{\begin{tabular}[c]{@{}c@{}}Rot.\\ error ($^\circ$)\end{tabular}} & \multicolumn{1}{c|}{\begin{tabular}[c]{@{}c@{}}Trans.\\ error (m)\end{tabular}} & \multicolumn{1}{c|}{\begin{tabular}[c]{@{}c@{}}Rot.\\ error ($^\circ$)\end{tabular}} \\ \hline
	Odometry                                          & 0.96$\pm$0.43                                                                       & 10.01$\pm$4.54                                                                    & 0.79$\pm$0.32                                                                       & 6.27$\pm$3.54                                                                     & 0.62$\pm$0.23                                                                       & 3.71$\pm$2.38                                                                     \\ \hline
	Pose est. with ranging (Sect. III-B)              & 0.48$\pm$0.26                                                                       & 7.57$\pm$5.57                                                                     & 0.39$\pm$0.24                                                                       & 5.27$\pm$4.34                                                                     & 0.37$\pm$0.22                                                                       & 5.05$\pm$3.63                                                                     \\ \hline
	Pose opt. with odom. (w=5)           & 0.42$\pm$0.18                                                                       & 7.02$\pm$5.53                                                                     & 0.35$\pm$0.17                                                                       & 4.77$\pm$4.15                                                                     & 0.33$\pm$0.18                                                                       & 4.69$\pm$3.95                                                                     \\ \hline
	Pose opt. with odom. (w=30)           & 0.38$\pm$0.14                                                                      & 4.46$\pm$2.24                                                                     & 0.33$\pm$0.13                                                                       & 3.75$\pm$2.36                                                                     & 0.32$\pm$0.13                                                                       & 3.19$\pm$2.63                                                                     \\ \hline
	Pose opt. with odom. (w=80)          & 0.33$\pm$0.11                                                                       & 2.99$\pm$2.33                                                                     & 0.30$\pm$0.10                                                                       & 2.55$\pm$2.06                                                                     & 0.30$\pm$0.12                                                                       & 2.35$\pm$2.56                                                                     \\ \hline
	Pose opt. with odom. (w=160)         & 0.37$\pm$0.18                                                                       & 3.49$\pm$1.86                                                                     & 0.30$\pm$0.14                                                                       & 2.91$\pm$1.76                                                                     & 0.31$\pm$0.15                                                                       & 3.09$\pm$2.34                                                                     \\ \hline
	PF with UWB rangings \cite{ref10}                             & 0.35$\pm$0.20                                                                       & 5.31$\pm$3.91                                                                     & 0.49$\pm$0.23                                                                       & 6.26$\pm$3.84                                                                     & 0.35$\pm$0.22                                                                       & 4.84$\pm$3.12                                                                     \\ \hline
	PF with optimized pose (Sect. III-D)              & 0.25$\pm$0.10                                                                       & 2.91$\pm$2.76                                                                     & 0.22$\pm$0.10                                                                       & 2.70$\pm$2.27                                                                     & 0.25$\pm$0.09                                                                       & 2.02$\pm$1.47                                                                     \\ \hline
\end{tabular}
\end{table*}

\subsection{First Test Case: One Static and One Mobile Robot}
In this test case, one robot is placed at the area and remains static. Another robot is controlled to move along a rectangular path several times.
The goal of this series of experiments is to evaluate the feasibility of the proposed approach. These experiments also help us to examine the impact of different parameters on the localization accuracy. Since the distance between the UWB nodes on a robot has high impact on pose estimation, we tested the following three distance configurations of the UWB nodes, namely 0.3m, 0.5m, and 0.7m. 

A summary of the positioning results with different approaches and different UWB configurations are shown in Table \ref{table_static}. 
We also compared our approach with the traditional approach that uses the UWB ranging to update the particle weights \cite{ref10}. As can be seen from Table\,\ref{table_static}, with a large distance setting of the UWB, we obtain a slightly better localization accuracy. The table also shows that pose optimization with additional odometry constraints gives improvement to the localization accuracy. Regarding the sliding window size $w$, a large sliding window produces a better localization accuracy. 
On the other hand, optimizing with a large sliding window requires more computational time. 

As it can be also seen from Table \ref{table_static}, our particle filtering, which fuses the odometry and the optimized pose (with a sliding window $w=30$), gives the best localization accuracy. Figure \ref{two_robots_track}(a) shows the path estimated with different approaches. 
Figure \ref{two_robots_track}(b) and Figure \ref{two_robots_track}(c) plot the localization accuracy in translation and rotation with respect to different timestamps.  

\subsection{Second Test Case: Three Mobile Robots}

\begin{figure*}
 \subfigure[]{

    \includegraphics[width=2.2in]{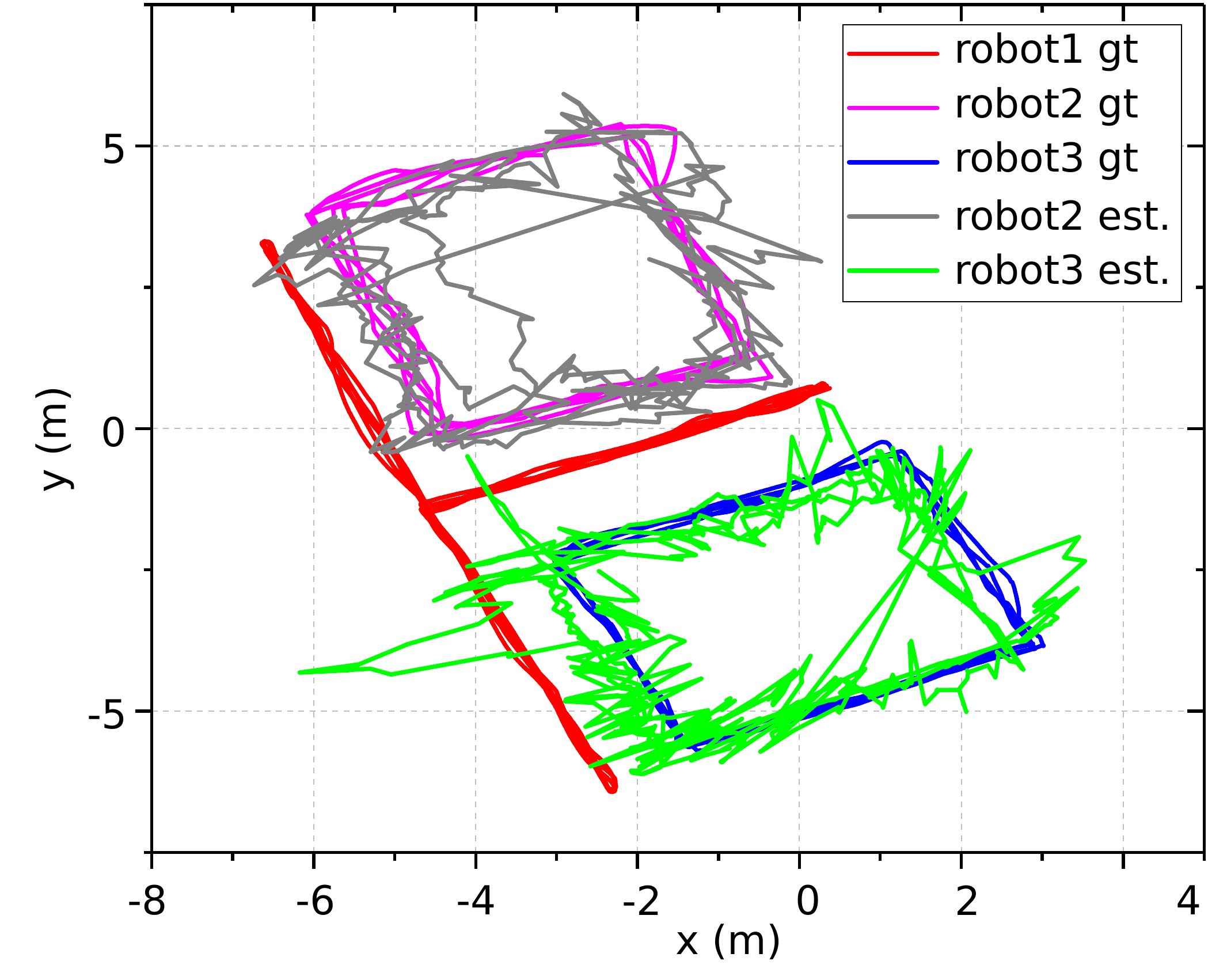} 

}
\subfigure[]{


    \includegraphics[width=2.2in]{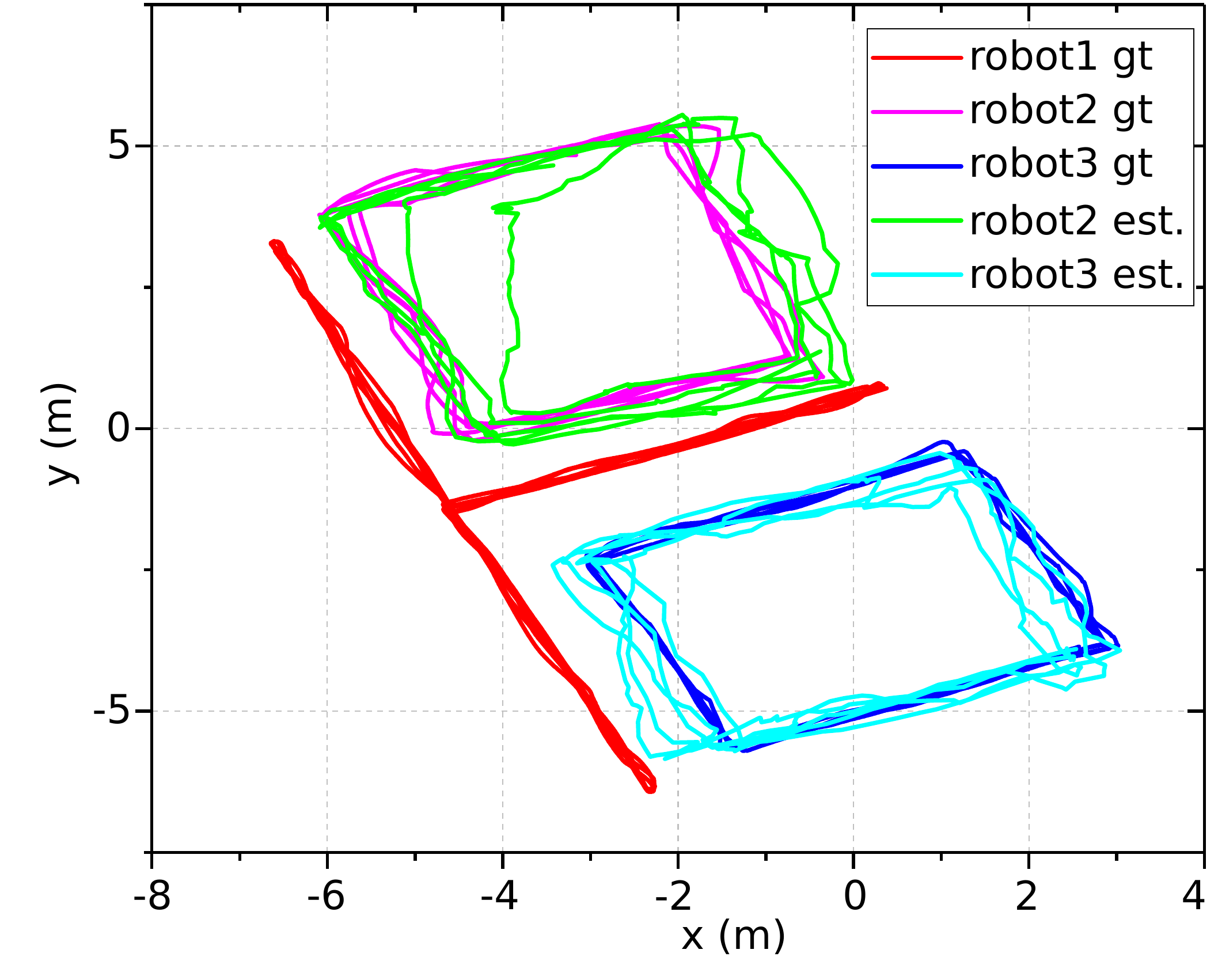}

}
\subfigure[]{


    \includegraphics[width=2.2in]{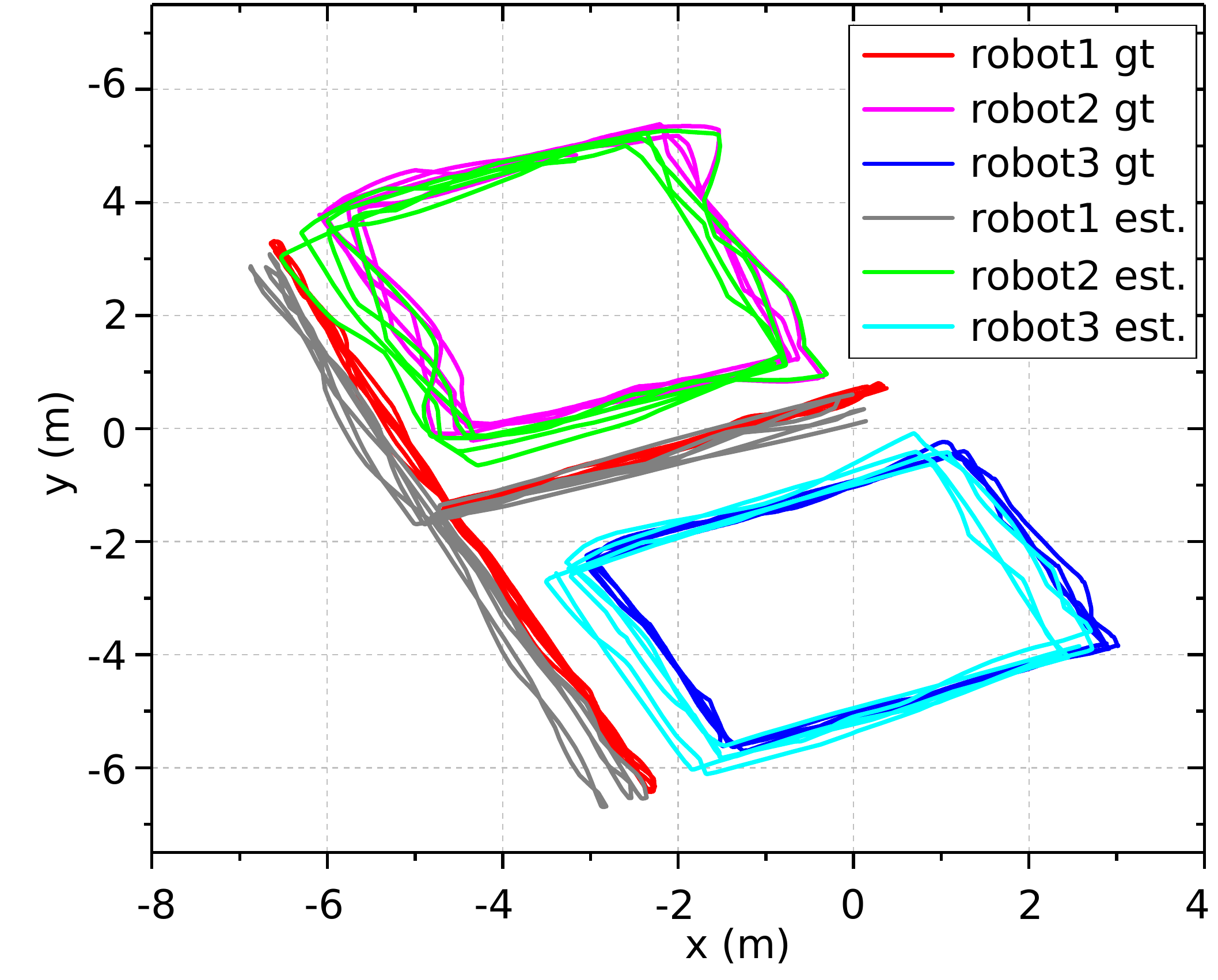} 

}

\caption{Trajectories estimated by different approaches. (a) Relative pose estimation of robot2 and robot3 by UWB ranging (Section\,\ref{pose_est_ranging}) assuming known robot1 position; (b) Optimized pose of robot2 and robot3 with additional odometry constraints (Section\,\ref{pose_est_g2o}) assuming known robot1 position;  (c) The particle filtering by fusing odometry and optimized pose (Section\,\ref{pose_est_pf}).}

\label{three_robot_track} 

\end{figure*}

To better verify the feasibility of our proposed approach, we controlled three robots to move along different paths.
Figure\,\ref{three_robot_track} shows the trajectory estimated by different approaches. 
We show the relative positioning error between the three robots under different approaches in Table\,\ref{table_three}.
We set the sliding window $w=30$ in this set of experiments. 
As can be see from Table\,\ref{table_three}, the odometry shows an accumulative error of 1.466m in translation and 22.076$^\circ $ in rotation. The use of UWB obviously produces an improvement of the localization accuracy.
The pure UWB ranging based-approach (i.e., pose estimation based on pure UWB ranging in Section \ref{pose_est_ranging}) provides a translational error of 0.528m and rotational error of 7.413$^\circ$, while this accuracy is improved to 0.312m in translation and 4.903$^\circ$ in rotation by the particle filtering that integrates odometry and optimized pose obtained with a sliding window.
 In general, the accuracy we obtained with three robots is much worse than the first test case, due to the increase of the experimental space and the UWB ranging error cause by the moving of robots.

Table \ref{table_three} also compares the computational time of different approaches for the second test case. 
We ran the algorithm on a laptop with an Intel i5-6300HQ 2.30GHz CPU and 12.0G RAM.
In total, the proposed approach consumes approx. 81ms to perform one sensor update, 
including relative pose estimation in Section \ref{pose_est_ranging} (21 ms), 
pose optimization by sliding window with $w=$ 30 in Section \ref{pose_est_g2o} (41 ms), and the sensor fusion by the particle filtering with a particle size of 500 in Section \ref{pose_est_pf} (45 ms). 
This allows us to produce the estimation at a frequency of 12Hz (i.e., 1000/81$\approx$12), which is suitable for many robotics applications. 

\begin{table}[]
\centering
\caption{Evaluation of the second test case: the translational (in meters) and rotational error (in degrees) between three robots and average computational time (in millisecond) for one pose update under different approaches.}

\label{table_three}
\begin{tabular}{|c|c|c|c|}
\hline
Approaches                                                                                           & \begin{tabular}[c]{@{}c@{}}Trans.\\error (m) \end{tabular} & \begin{tabular}[c]{@{}c@{}}Rot.\\error ($^\circ$)\end{tabular} & \begin{tabular}[c]{@{}c@{}}Consumed\\time (ms)\end{tabular} \\ \hline
Odometry                                                                                       & 1.466                                                  & 22.076                                                &          ---                                                  \\ \hline
\begin{tabular}[c]{@{}c@{}}Pose estimation\\ from UWB ranging\end{tabular}              & 0.528                                                  & 7.413                                                & 21                                                           \\ \hline
\begin{tabular}[c]{@{}c@{}}Pose optimization\\ with odom. ($w$=30)\end{tabular}                     & 0.424                                                  & 5.648                                                 & 41                                                           \\ \hline
\begin{tabular}[c]{@{}c@{}}PF with\\UWB rangings\end{tabular}                         & 0.534                                                  & 7.234                                                 & 44                                                           \\ \hline
\begin{tabular}[c]{@{}c@{}}PF with\\optimized pose\end{tabular} & 0.312                                                  & 4.903                                                 & 86                                                           \\ \hline
\end{tabular}

\end{table}

\section{conclusion}
We proposed an approach for relative localization without any infrastructure based on the fusion of multiple UWB ranging measurements from different nodes installed on the robots. 
We also proposed to optimize the pose estimation by incorporating additional odometry constraints. 
This optimized pose is then fused with a particle filtering for the pose tracking of a group of robots. 
Our experiments show that, with three robots moving in the environment, we are able to achieve an accuracy of 0.528m in translation and 7.413$^\circ$ in rotation (in an environment with a size of 6m$\times$12m) based on pure UWB ranging measurements. 
The accuracy is improved to 0.424m in translation and 5.648$^\circ$ in rotation by adding odometry constraints for optimization. 
The particle filtering that fuses odometry and optimized pose provides a translational error of 0.312m and rotational error of 4.903$^\circ$. 
Our approach provides a solution for the localization of a team of robots without any knowledge about the infrastructure. 
In the future work, we would like to extend our work to include more robots and apply our approach for swarm and formation control of multiple robots.

\end{document}